%% file: bmvc_final.tex
\documentclass{bmvc2k}
\usepackage{multirow}

\title{Unlocking Spatial Grounding in Flow-Matching Models via SNR Tuning}

\addauthor{Martin Thißen}{martin.thissen@h-da.de}{1, 2}
\addauthor{Elke Hergenröther}{elke.hergenroether@h-da.de}{1, 2}

\addinstitution{
   Department of Computer Science \\
   Darmstadt University of Applied Sciences \\
   Darmstadt, Germany
}

\runninghead{Thißen, Hergenröther}{Unlocking Spatial Grounding via SNR Tuning}

\def\etal{\emph{et al}\bmvaOneDot}

\usepackage{float}

\begin{document}

\maketitle

\input{chapters/abstract.tex}
\input{chapters/introduction.tex}
\input{chapters/related_work.tex}
\input{chapters/method.tex}
\input{chapters/results.tex}
\input{chapters/conclusion.tex}

\raggedbottom
\bibliography{egbib}

\pagebreak
\appendix
\label{appendix:firstpage}
\input{chapters/appendix.tex}
\end{document}

%% file: chapters/abstract.tex
\begin{abstract}
\noindent
Despite the success of foundation models in image synthesis, achieving fine-grained spatial control remains a significant challenge, typically requiring auxiliary structural adapters or complex inference-time steering. While modern vision-language (VLM) text encoders possess the latent capacity to represent spatial coordinates, generative models often fail to leverage this information. We demonstrate that the signal-to-noise ratio (SNR) distribution is a critical determinant of spatial grounding in flow-matching models. Using a novel synthetic benchmark of geometric primitives, we isolate structural mechanics from textural complexity inherent in natural imagery and introduce a rigorous color-aware layout metric to quantify spatial fidelity. We show that training on a centered SNR distribution results in poor layout adherence (25.9\%). Instead, we identify high-noise regimes as a "critical learning phase" for spatial grounding. Shifting the training focus toward these stages increases layout adherence to 80.8\%, effectively reaching the 83.6\% reconstruction ceiling of the underlying variational autoencoder. We find that the necessity for this shift scales with resolution and aspect ratio variance, and that the conservative noise schedules employed by models like FLUX.1 are insufficient for precise layout adherence. Finally, we show that these findings translate to real-world imagery via the COCO dataset (1.0 vs. 57.3 Average Precision), where state-of-the-art layout adherence is unlocked through distribution tuning alone, without architectural modifications.
\end{abstract}

%% file: chapters/introduction.tex
\section{Introduction}
\label{sec:intro}
The rapid evolution of diffusion \cite{ho2020denoising, song2021score} and flow-matching \cite{lipman2023flow, liu2023flow} models is outpacing the adoption of these models for specialized use cases. In particular, while new state-of-the-art image synthesis models \cite{2025qwenimage,team2025zimage, flux-2-2025, cao2025hunyuanimage} are frequently released, achieving fine-grained spatial control over their outputs typically requires adopting one of two paradigms. The first involves training-based methods, which utilize auxiliary structural adapters such as ControlNet \cite{Zhang_2023_ICCV}, GLIGEN \cite{Li_GLIGEN_2023_CVPR}, or CreatiLayout \cite{zhang2025creatilayout}. The second relies on training-free methods, which steer the generation process of frozen backbones during inference to follow a target layout. However, adapters are often tightly coupled to specific model architectures, making them costly to retrain for each new release, while training-free methods often struggle with complex layouts or specific architectural constraints. This creates a critical need to understand why foundation models struggle to generate images that follow layouts from text alone.\\

\noindent Standard pre-trained models typically fail to interpret 2D coordinates embedded in text prompts, often treating them as non-spatial semantic tokens rather than geometric constraints. In contrast, recent advancements in vision-language models (VLMs), such as the Qwen-VL series \cite{Qwen2.5-VL, Qwen3-VL}, demonstrate that multimodal embeddings can successfully encode spatial relationships and discrete coordinates for tasks like object detection and grounding. This suggests that their multimodal embedding space already possesses the latent capacity to represent spatial information. As state-of-the-art text-to-image models increasingly adopt VLM-based text encoders, it remains an open question why the generative process fails to leverage this innate grounding capability. Therefore, we investigate whether, and under what conditions, text-to-image models can learn to follow explicit 2D layout specifications encoded solely in text prompts.\\

\begin{figure}[bth]
\centering
  \subfigure[Synthetic Images]{
    \includegraphics[width=0.48\textwidth]{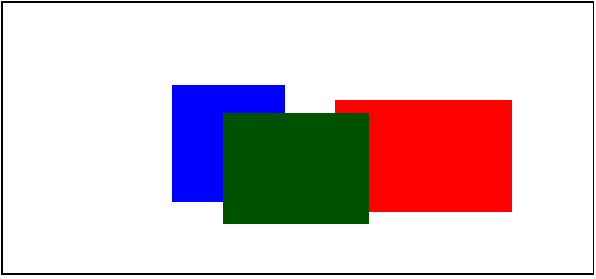}
    \label{fig:example_synthetic_dataset_intro}
  }
  \hfill
  \subfigure[Real-World Images]{
    \includegraphics[width=0.48\textwidth]{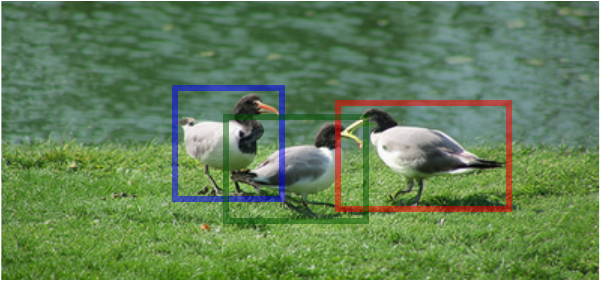}
    \label{fig:example_real_world_dataset_intro}
  }
\caption{\textbf{From Synthetic Experiments to Real-World Transfer.} Layout adherence is first studied in a controlled synthetic setting, using color-filled regions as exact ground truth (left). This allows an isolated validation before extending the approach to real images (right).}
\label{fig:example_synthetic_real_dataset}
\end{figure}

\noindent To this end, we examine the grounding capabilities of flow-matching models by fine-tuning them to interpret 2D coordinates directly from text prompts without introducing structural modifications. To isolate the mechanisms of spatial grounding, we first construct a controlled synthetic dataset of geometric primitives, enabling exact, pixel-level assessment of layout fidelity (Fig.~\ref{fig:example_synthetic_dataset_intro}). Our analysis reveals that the signal-to-noise ratio (SNR) distribution is the critical determinant of spatial grounding. We find that training on a balanced SNR distribution actually hinders the model from accurately learning coordinate-based constraints. Instead, biasing the training toward high-noise regimes, where global structure is established, allows the model to internalize the relationship between text tokens and spatial coordinates. Notably, we find that this distribution shift is necessary even at lower resolutions ($256^2$), and that the conservative shifts employed by recent models like FLUX.1~\cite{labs2025flux1kontextflowmatching} are insufficient for precise layout adherence. Finally, we demonstrate that these principles successfully translate to real-world imagery via the COCO dataset~\cite{lin2014microsoft} (Fig.~\ref{fig:example_real_world_dataset_intro}), showing that state-of-the-art layout adherence can be unlocked in standard flow-matching models through distribution tuning alone, without the need for specialized architectural extensions.\\

\noindent Taken together, our results demonstrate that high-fidelity spatial control is a latent capability in state-of-the-art flow-matching models that can be unlocked by strategically shifting the SNR distribution during training, without requiring any architectural modifications. We show that the high-noise regime serves as a "critical learning phase" for spatial grounding, forcing the model to internalize geometric constraints when global structure is not yet observable. Our main contributions are as follows:
\begin{itemize}
\item \textbf{Identification of SNR as a Key Facilitator:} We provide a systematic study showing that biasing the SNR distribution toward high-noise regimes is essential for teaching models to interpret 2D coordinates embedded in text.
\item \textbf{A Synthetic Benchmark for Spatial Precision:} We introduce a benchmark consisting of geometric primitives that enables the exact, pixel-level quantification of layout adherence, isolating grounding mechanics from the textural complexities of real-world images.
\item \textbf{Transfer to Real-World Imagery:} We provide empirical evidence that these insights translate to complex real-world imagery via the COCO dataset, achieving state-of-the-art layout adherence in standard flow-matching models through distribution tuning alone, bypassing the need for specialized structural adapters.
\end{itemize}

%% file: chapters/related_work.tex
\section{Related Work}
\label{sec:related_work}
\textbf{Image Synthesis.}
Generative modeling has transitioned from Generative Adversarial Networks (GANs)~\cite{NIPS2014_f033ed80} to Diffusion \cite{ho2020denoising, song2021score, dhariwal2021diffbeatsgans} and Flow Matching \cite{lipman2023flow, liu2023flow} frameworks. Architectures have similarly evolved from convolutional U-Nets (e.g., Stable Diffusion \cite{Rombach_2022_CVPR}, SDXL \cite{podell2024sdxl}) toward scalable Diffusion Transformers (DiT) \cite{Peebles_2023_ICCV} and Multimodal Diffusion Transformers (MMDiT) \cite{esser_stable_diffusion_3, flux2024, labs2025flux1kontextflowmatching}. Recent models \cite{2025qwenimage, flux-2-2025} replace traditional text encoders (e.g., CLIP \cite{Radford2021CLIP} or T5 \cite{Raffel_T5}) with Vision-Language Models (VLMs) \cite{liu2026ministral3, Qwen2.5-VL, Qwen3-VL} to leverage pre-aligned latent spaces. However, while VLMs excel at spatial grounding tasks like detection \cite{Qwen2.5-VL, Qwen3-VL}, generative models using them still struggle to natively interpret 2D coordinates within text prompts, necessitating external spatial control.

\noindent
\textbf{Training Dynamics and SNR Schedules.} 
The performance of flow-matching and diffusion models heavily depends on the timestep sampling distribution $p(t)$, which determines the Signal-to-Noise Ratio (SNR) levels encountered during training \cite{kingma2023understanding, esser_stable_diffusion_3, bfl2025representation}. While logit-normal distributions are typically tuned to improve visual fidelity based on resolution \cite{hoogeboom2023simple, labs2025flux1kontextflowmatching, esser_stable_diffusion_3} or dimensionality \cite{zheng2026diffusion}, their impact on spatial grounding is only beginning to be examined. Recent work by Wu \etal~\cite{wu2025hybrid} and Zhang \etal~\cite{zhang2025creatilayout} provides evidence that noise schedules influence layout adherence. However, their findings are coupled with specialized regional or siamese architectures. In contrast, our work demonstrates that the SNR distribution alone serves as the primary lever for internalizing spatial-semantic mappings in standard, structurally unmodified flow-matching models. We show that the acquisition of spatial grounding from text prompts alone is highly sensitive to $p(t)$, revealing that high-noise regimes represent the essential window for anchoring textual coordinates to global structural composition.

\noindent
\textbf{Layout-to-Image (L2I) Generation.} Layout-to-image approaches address the limitations of native spatial control by augmenting pre-trained synthesis models with explicit geometric conditioning (e.g., bounding boxes or masks). This is increasingly viewed as a multi-stage pipeline, beginning with upstream planning where Large Language Models (LLMs) or autoregressive VLMs translate linguistic descriptions into structural layouts \cite{ske_layout, he2025plangen, Feng_2024_CVPR}. Other approaches bridge the gap between text and pixels by synthesizing intermediate representations, such as depth maps, to guide the final generative process \cite{zhou2025dis, zhou20253disflux, 2025buildascene}. Once a spatial structure is established, the literature diverges into training-based and training-free strategies. Training-based L2I methods adapt model architectures to interpret spatial signals via fine-tuning or dedicated adapters \cite{Li_GLIGEN_2023_CVPR, Zheng_2023_CVPR, Yang_2023_ICCV}. These works have evolved to handle multi-instance scenarios by using spatial attention disentanglement \cite{Zhou_2024_CVPR, Wang_2024_CVPR} or hierarchical composition \cite{bocheng2024hico} to minimize attribute leakage. Beyond image quality, these models often serve as data engines to synthesize paired training data for perception tasks \cite{feng2024instagen, zhu2024odgen}. Recently, MMDiT architectures have been enhanced with siamese attention to isolate layout signals \cite{zhang2025creatilayout} and hybrid control frameworks to achieve data-efficient grounding \cite{wu2025hybrid}. To avoid the high cost of retraining, training-free L2I methods intervene in the sampling process of frozen backbones. These strategies primarily manipulate internal cross-attention maps to ensure semantic isolation of subjects \cite{xue2023freestylenet, boundedattention2024, zhan2025larender} or employ optimization-based guidance to steer the denoising trajectory toward specified coordinates \cite{layoutguidance2024, attre2024, xiao2024rb, csg2024, winwinlay2025}. Beyond attention manipulation, compositional strategies decentralize the process by merging localized per-object noise estimates or synchronizing global and instance-specific denoising branches \cite{noisecollage2024, groundit2024}. Finally, probabilistic approaches like Psi-Sampler \cite{psisampler2025} reinterpret layout adherence as an inference-time reward alignment problem, using Sequential Monte Carlo sampling to filter trajectories.\\

\noindent
While existing L2I methods typically rely on external modules to bypass spatial reasoning limitations, this study investigates a more fundamental question: can flow-matching models acquire spatial grounding from text prompts alone, without any structural modifications?

%% file: chapters/method.tex
\section{Method}
\label{sec:method}
\subsection{Synthetic Benchmark for Text-Layout Image Generation}
To facilitate training and provide a controlled environment for evaluation, we construct a synthetic dataset comprising 100,000 training pairs. For rigorous quantitative evaluation, we hold out a curated benchmark set of 1,000 text-image pairs. Images are rendered at resolution $W \times H$ and contain $n \sim \mathcal{U}\{1, 10\}$ rectangles. In our standard benchmark, we set $W = H = 256$. Notably, our generation pipeline is agnostic to resolution and aspect ratio, enabling us to evaluate scalability across higher resolutions (e.g., $1024^2$) and varying aspect ratios, as detailed in Section~\ref{sec:results_aspect_ratio_resolution}. By utilizing simple geometric primitives, this approach mitigates the textural complexity and photometric noise present in real-world images. This simplification enables us to disentangle spatial control from semantic rendering, allowing for a systematic and quantitative evaluation of a model's geometric layout compliance.

\noindent
\textbf{Layout Synthesis.} We adopt rectangles as spatial primitives to mirror standard bounding-box formats and enable precise evaluation. To facilitate a systematic study of spatial grounding, we sample geometric properties from an extensive distribution that spans diverse scales and aspect ratios. Specifically, we define aspect ratio $r$ and area $A$ through log-uniform sampling: $\ln(r) \sim \mathcal{U}(\ln \frac{1}{20}, \ln 20)$ and $\ln(A) \sim \mathcal{U}(\ln A_{min}, \ln A_{max})$, where $A_{min} = 0.05W \times 0.05H$ and $A_{max} = 0.9W \times 0.9H$. Dimensions are then derived as $w = \sqrt{A \cdot r}$ and $h = \sqrt{A / r}$. This diverse geometric sampling ensures the benchmark provides a robust evaluation of spatial control, testing the model's ability to generalize across a wide range of object shapes.

\noindent
\textbf{Scene Composition.} We sample centers $(c_x, c_y)$ uniformly and clip the resulting rectangles to the image frame $[0, W] \times [0, H]$. As rectangles are rendered sequentially, we remove fully obscured rectangles from the metadata and prompt to ensure semantic consistency. This guarantees that all described objects are at least partially visible, preventing a mismatch between the textual description and visual ground truth.

\noindent
\textbf{Semantic Mapping.} Each rectangle is assigned a unique color from a fixed palette of ten distinct values. To ensure maximum discriminability, which is important for our layout adherence metric introduced in Section~\ref{subsec:layout_adherence_metric}, these colors are selected to be maximally distant from one another in the CIELAB color space. Each color is mapped to its corresponding linguistic label using the XKCD color naming convention \cite{xkcd_color_survey} (e.g., fluro green).

\begin{figure}[t]
\centering
  \subfigure[Text Prompt]{
    \includegraphics[width=0.38\textwidth]{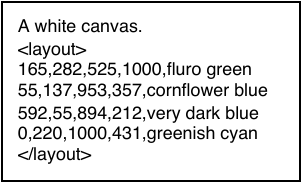}
    \label{fig:example_synthetic_dataset_sub_1}
  }
  \hfill
  \subfigure[Underlying Occlusion]{
    \includegraphics[width=0.26\textwidth]{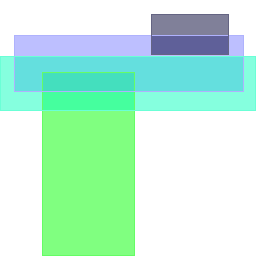}
    \label{fig:example_synthetic_dataset_sub_2}
  }
  \hfill
  \subfigure[Final Image]{
    \includegraphics[width=0.26\textwidth]{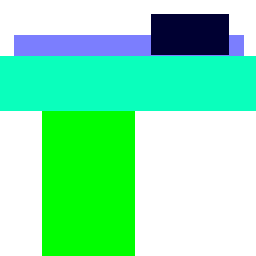}
    \label{fig:example_synthetic_dataset_sub_3}
  }
\caption{\textbf{Example $\langle \textit{text prompt}, \textit{image} \rangle$ pair from our synthetic dataset:} (a) the input text prompt used to condition the scene; (b) a visualization of the occlusion, where rectangles are rendered with semi-transparency to reveal depth ordering; and (c) the final rendered image.}
\label{fig:example_synthetic_dataset}
\end{figure}

\noindent
\textbf{Layout Encoding.} Following Bai \etal~\cite{Qwen3-VL}, we normalize coordinates to $[0, 1000]$. A typical object is formatted as: $[x_{min}, y_{min}, x_{max}, y_{max}, \text{color}]$. Crucially, rectangles are listed in the prompt in their exact rendering order, establishing a spatial hierarchy where the textual sequence dictates occlusion priority. To succeed, the model must not only solve the challenging problem of spatial grounding but also master the compositional reasoning required to render overlapping geometries where objects are frequently partially obscured. Figure~\ref{fig:example_synthetic_dataset} visualizes this hierarchy across the prompt, occlusion layers, and final rendered image.

\subsection{Layout Adherence Metric}
\label{subsec:layout_adherence_metric}
To assess the spatial fidelity of generated images relative to specified text layouts, we introduce a Layout Adherence Metric based on color-coded segmentation masks. This metric quantifies the alignment between the ground-truth image $I_{\text{gt}}$ and synthesized output $I_{\text{gen}}$ by first transforming both into the perceptually uniform CIELAB color space. We then identify the set of unique foreground colors $\mathcal{C}$ present in $I_{\text{gt}}$, where $|\mathcal{C}| \leq 10$ corresponds to the number of layout rectangles. For each color $c \in \mathcal{C}$, we derive binary masks for both the ground-truth and generated images. Specifically, binary masks $M_c(I; \tau)$ are derived for both images by thresholding the perceptual color distance between each pixel $\mathbf{p}$ and the target color $c$:
\begin{equation}
    M_c(I; \tau) = \left\{ \mathbf{p} \in \Omega \mid \| I(\mathbf{p}) - c \|_2 \leq \tau \right\},
\end{equation}
where $\Omega$ denotes the image spatial domain and $\tau$ is a predefined color-matching threshold. This masking process accounts for minor color deviations in the generative process while isolating the spatial extent of each layout element. The adherence for a specific color $c$ is then calculated using the Intersection over Union ($\mathrm{IoU}$) between the respective masks:
\begin{equation}
\label{eq:iou_c}
    \mathrm{IoU}_{c}(I_{\mathrm{gt}}, I_{\mathrm{gen}}; \tau) = \frac{|M_{c}(I_{\mathrm{gt}}; \tau) \cap M_{c}(I_{\mathrm{gen}}; \tau)|}{|M_{c}(I_{\mathrm{gt}}; \tau) \cup M_{c}(I_{\mathrm{gen}}; \tau)|}.
\end{equation}

\begin{figure}[t]
\centering
  \subfigure[Ground Truth]{
    \includegraphics[width=0.18\textwidth]{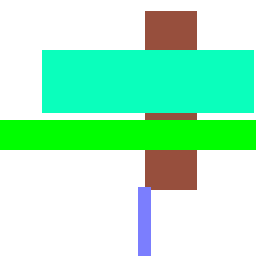}
    \label{fig:miou_img_1}
  }
  \subfigure[\parbox{0.1\textwidth}{\centering Darkened \\ \scriptsize $\mathrm{mIoU}_{10}\!=\!0$ \\ $\mathrm{mIoU}_{15}\!=\!1$}]{
    \includegraphics[width=0.18\textwidth]{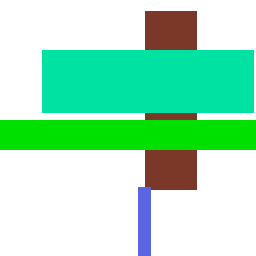}
    \label{fig:miou_img_2}
  }
  \subfigure[\parbox{0.13\textwidth}{\centering Misaligned \\ \scriptsize $\mathrm{mIoU}_{2:25} \approx 0.52$}]{
    \includegraphics[width=0.18\textwidth]{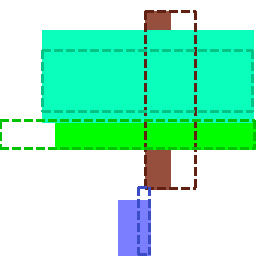}
    \label{fig:miou_img_3}
  }
  \subfigure[\parbox{0.14\textwidth}{\centering Missing Object \\ \scriptsize $\mathrm{mIoU}_{2:25} = 0.75$}]{
    \includegraphics[width=0.18\textwidth]{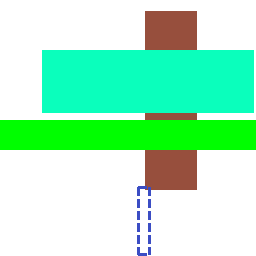}
    \label{fig:miou_img_4}
  }
  \subfigure[\parbox{0.14\textwidth}{\centering Reversed Order \\ \scriptsize $\mathrm{mIoU}_{2:25} \approx 0.75$}]{
    \includegraphics[width=0.18\textwidth]{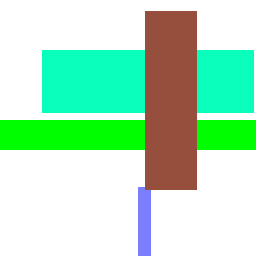}
    \label{fig:miou_img_5}
  }
  \caption{\textbf{Layout Adherence Metric Sensitivity.} We illustrate $\mathrm{mIoU}_\tau$ across different failure modes. The metric effectively penalizes spatial misalignment (c), missing elements (d), and incorrect occlusion ordering (e). Note that spatial misalignment can result in a lower score than an omission, reflecting the metric's focus on precise spatial grounding.}
  \label{fig:miou_comparison}
\end{figure}

\noindent
Finally, we compute the mean IoU ($\mathrm{mIoU}_\tau$) across all unique colors in the image:
\begin{equation}
\label{eq:miou_tau}
    \mathrm{mIoU}_\tau = \frac{1}{|\mathcal{C}|} \sum_{c \in \mathcal{C}} \mathrm{IoU}_{c}(I_{\mathrm{gt}}, I_{\mathrm{gen}}; \tau).
\end{equation}
As illustrated in Fig.~\ref{fig:miou_comparison}, this metric effectively penalizes both spatial misalignments and the omission of required layout elements. To ensure the metric is not overly sensitive to the choice of the threshold $\tau$ (Fig.~\ref{fig:miou_img_2}), we report the mean performance over a discrete set of thresholds $\mathcal{T} = \{2, 3, \dots, 25\}$, defined as:

\begin{equation}
\label{eq:miou_2_25}
\text{mIoU}_{2:25} = \frac{1}{|\mathcal{T}|} \sum_{\tau \in \mathcal{T}} \mathrm{mIoU}_\tau.
\end{equation}

\noindent
This allows us to observe the trade-off between strict color matching and spatial alignment. Furthermore, we establish an empirical upper bound for this metric by calculating the adherence of the VAE-reconstructed ground truth ($I_{\text{vae}} = \mathcal{D}(\mathcal{E}(I_{\text{gt}}))$). This serves as a calibration point to isolate generative errors from the autoencoder's reconstruction limitations.

\subsection{Implementation Details}
\textbf{Fine-Tuning.} While our primary experiments utilize FLUX.2-4B-base, we also scale our approach to larger models including FLUX.2-9B-base \cite{flux-2-2025}, FLUX.1-dev (12B) \cite{labs2025flux1kontextflowmatching,flux2024}, and Qwen-Image (20B) \cite{2025qwenimage}. To enable fine-tuning on a single 80GB A100 GPU, we utilize the Adafactor optimizer \cite{shazeer2018adafactor} (with constant scale, disabling relative step and parameter scaling) alongside a fused backward pass and gradient checkpointing. This configuration, combined with BFloat16 precision, reduces the peak memory footprint of the 20B Qwen-Image model from 160GB to 44GB at $256 \times 256$ resolution. We fine-tune for 100,000 steps using a unit batch size, as layer-wise fused updates are mathematically incompatible with gradient accumulation. This approach leverages recent findings that small-batch training maintains competitive convergence and generalization for large generative models \cite{marek2026small}. To ensure training stability and preserve generation fidelity, we employ an EMA model (0.9995 decay, $\gamma = 10$ Karras schedule \cite{Karras_2024_CVPR}) and a cosine learning rate schedule ($1\times10^{-5}$ to $5 \times 10^{-6}$) following a 5,000-step warmup. Finally, 10\% of prompts are nullified for classifier-free guidance (CFG).

\noindent
\textbf{Inference.} We generate images using 25 steps with a CFG scale of 3.0. We utilize a shifted logit-normal timestep scheduler ($\mu=1.0$), which provided the best empirical trade-off between speed and fidelity.

%% file: chapters/results.tex
\section{Experimental Results}
\label{sec:results}
Building on our setup, we systematically evaluate how SNR distribution, image resolution and aspect ratio, object properties, and model architecture influence the layout adherence of flow-matching models.

\subsection{Signal-to-Noise Ratio Distribution}
\label{sec:snr_dist}
\begin{figure}[t]
    \centerline{\includegraphics[width=\linewidth]{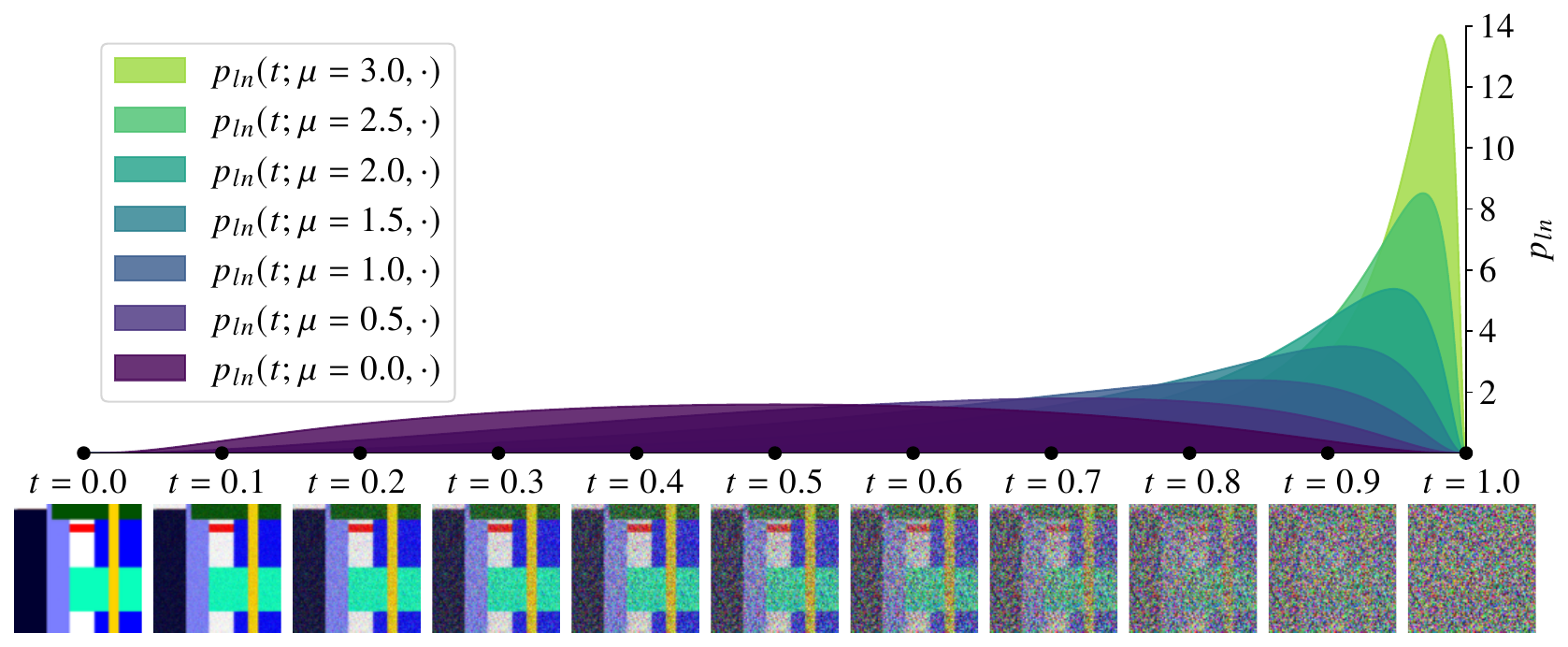}}
    \caption{\textbf{Shifted logit-normal distribution $\mathbf{p_{ln}}$ for location parameters $\mathbf{\mu}$.} Increasing $\mu$ shifts density toward higher noise levels (larger values of $t$), demonstrating how the distribution of training or sampling steps can be adjusted to prioritize specific denoising stages.}
    \label{fig:shifted_logit_normal_dist}
\end{figure}

\noindent
The SNR distribution during fine-tuning determines which noise levels the model encounters most frequently, directly influencing what it learns to denoise. This makes the timestep sampling distribution $t \sim p(t)$ a hyperparameter with significant impact on final model performance~\cite{kingma2023understanding, bfl2025representation}. In the context of rectified flow models \cite{liu2023flow}, the noised latent $x_t$ at timestep $t \in [0, 1]$ is defined by the interpolation:
\begin{equation}
    x_t = (1-t)x_0 + t\epsilon, \quad x_0 \sim p_{\text{data}}, \;\;\epsilon \sim \mathcal{N}(0, 1), \;\;t \sim p(t).
\end{equation}
We therefore investigate the effect of $p(t)$ on layout adherence and hypothesize that the model primarily learns the mapping between text coordinates and image locations during the high-noise stages, where it establishes global structure. Following recent advancements in flow-matching scaling~\cite{esser_stable_diffusion_3, hoogeboom2023simple, labs2025flux1kontextflowmatching, bfl2025representation}, we shift the logit-normal distribution:
\begin{equation}
    p_{ln}(t; \mu, \sigma) = \frac{1}{\sigma \sqrt{2\pi}} \cdot \frac{1}{t(1-t)} \exp\left(-\frac{(\mathrm{logit}(t) - \mu)^2}{2\sigma^2}\right),
\end{equation}
where $\mathrm{logit}(t) = \log \frac{t}{1-t}$. Following Batifol \etal~\cite{labs2025flux1kontextflowmatching,bfl2025representation}, we fix $\sigma = 1$ to isolate the effect of the mean SNR ($\mu$) shift on layout adherence. By adjusting $\mu$, we can systematically shift the fine-tuning focus toward the high-noise regime ($\mu > 0$), as illustrated in Fig.~\ref{fig:shifted_logit_normal_dist}. For our experiment, we investigate $\mu \in \left\{0.0, 0.5, 1.0, 1.5, 2.0, 2.5, 3.0 \right\}$.

\begin{figure}[t]
    \centerline{\includegraphics[width=\linewidth]{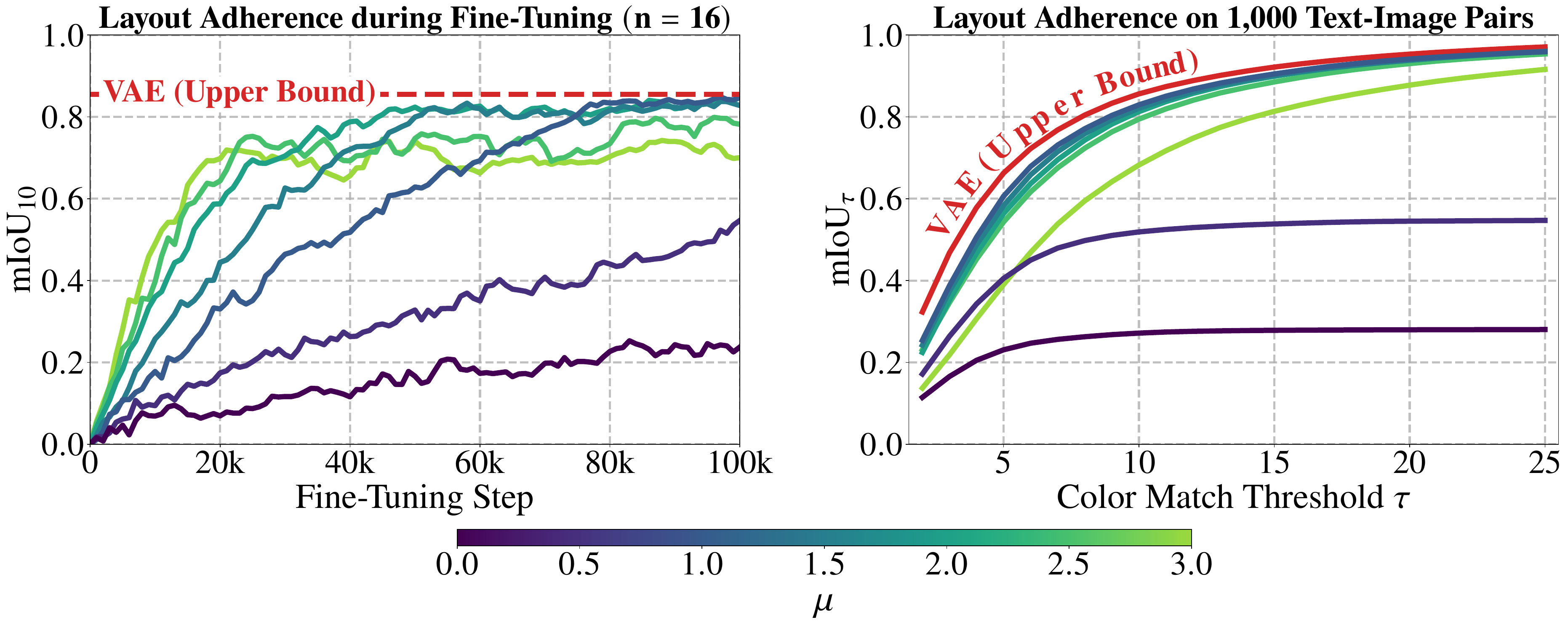}}
    \caption{\textbf{Layout adherence evaluation.} (Left) $\mathrm{mIoU}_{10}$ evolution over 100k steps for $n=16$ samples. (Right) Sensitivity to color match threshold $\tau$ across 1,000 text-image pairs. Higher $\mu$ values (except $\mu=3$) drive convergence toward the empirical VAE upper bound (red).}
    \label{fig:layout_adherence_experiment_4_1}
\end{figure}

\begin{figure}[t]
    \centerline{\includegraphics[width=\linewidth]{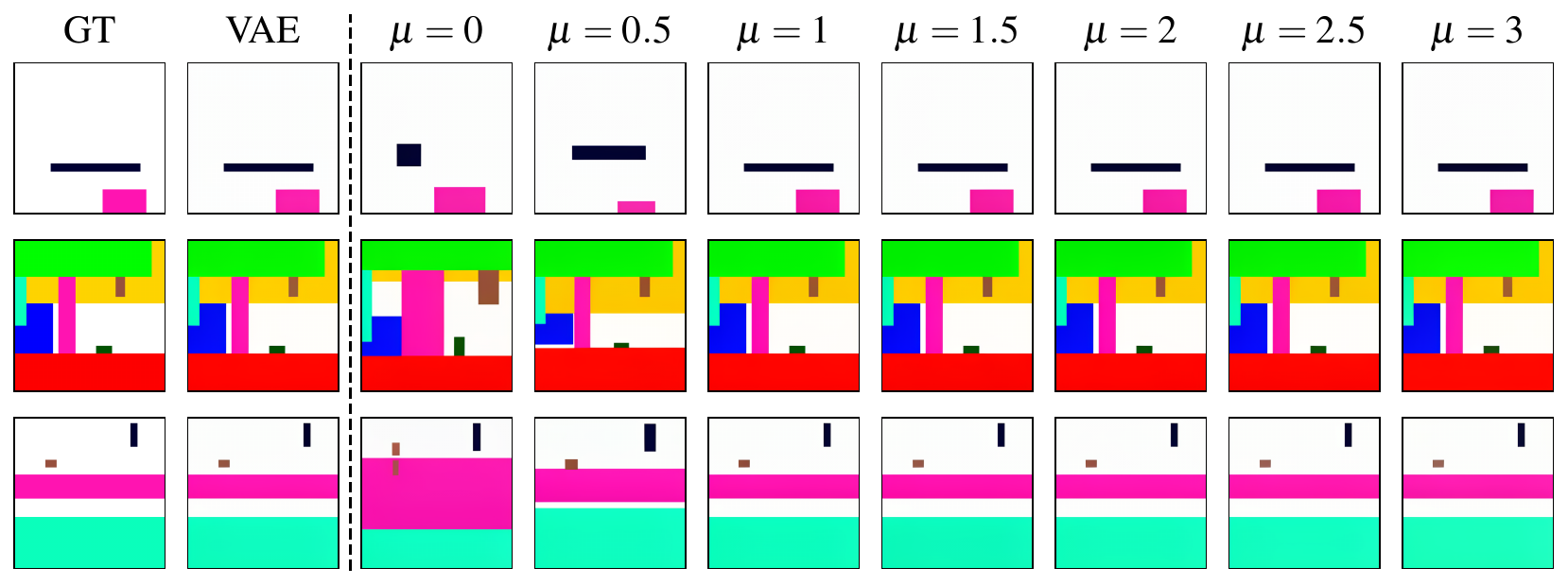}}
    \caption{\textbf{Qualitative comparison of generated images across varying $\mu$.} We compare the ground truth (GT) and VAE baseline against our model for $\mu \in [0, 3.0]$. Notably, spatial inaccuracies are easily perceivable for lower values ($\mu \in [0, 0.5]$). As $\mu$ increases, the generated images show noticeably higher fidelity and better spatial alignment with the GT.}
    \label{fig:qualitative_mu_sweep}
\end{figure}

\begin{table}[t]
    \centering
    \begin{minipage}[t]{0.48\linewidth}
        \centering
        \begin{tabular}{lccc}
            & \multicolumn{3}{c}{$\text{mIoU}_{2:25}$ $\uparrow$} \\
            \noalign{\smallskip}\cline{2-4}\noalign{\medskip}
            Method & $256^2$ & $512^2$ & $1024^2$ \\
            \noalign{\smallskip}\hline\noalign{\smallskip}
            $p_{ln}(t; \mu=3.0, \cdot)$           & 69.4          & 79.1           & 80.1 \\
            $p_{ln}(t; \mu=2.5, \cdot)$           & 78.1          & 79.8   & \textbf{82.7} \\
            $p_{ln}(t; \mu=2.0, \cdot)$           & 79.4& \textbf{83.8} & \underline{81.4} \\
            $p_{ln}(t; \mu=1.5, \cdot)$& \underline{80.1}& \underline{83.6}        & -- \\
            $p_{ln}(t; \mu=1.0, \cdot)$  & \textbf{80.8}          & 48.1           & -- \\
            $p_{ln}(t; \mu=0.5, \cdot)$           & 49.0          & --             & -- \\
            $p_{ln}(t; \mu=0.0, \cdot)$           & 25.9          & --             & -- \\
            \noalign{\smallskip}
            \hline
            \noalign{\smallskip}
            VAE (Recon.) \rule{0pt}{10pt}    & 83.6          & 87.6           & 88.3 \\
            \noalign{\smallskip}
            \hline
        \end{tabular}
        \caption{Effect of logit-normal shift $\mu$ on layout adherence across resolutions.}
        \label{tab:mu_ablation_resolution}
    \end{minipage}
    \hfill
    \begin{minipage}[t]{0.48\linewidth}
        \centering
        \begin{tabular}{lccc}
            & \multicolumn{3}{c}{$\text{mIoU}_{2:25}$ $\uparrow$} \\
            \noalign{\smallskip}\cline{2-4}\noalign{\medskip}
            Method & 1 & 2 & 3 \\
            \noalign{\smallskip}\hline\noalign{\smallskip}
            $p_{ln}(t; \mu=3.0, \cdot)$          & 69.4 & 73.5 & 69.3 \\
            $p_{ln}(t; \mu=2.5, \cdot)$           & 78.1 & \textbf{75.2} & \textbf{75.1} \\
            $p_{ln}(t; \mu=2.0, \cdot)$           & 79.4 & \textbf{75.2} & \underline{73.8} \\
            $p_{ln}(t; \mu=1.5, \cdot)$&\underline{80.1} & 74.4 & -- \\
            $p_{ln}(t; \mu=1.0, \cdot)$  & \textbf{80.8} & 67.8 & -- \\
            $p_{ln}(t; \mu=0.5, \cdot)$           & 49.0 & -- & -- \\
            $p_{ln}(t; \mu=0.0, \cdot)$           & 25.9 & -- & -- \\
            \noalign{\smallskip}
            \hline
            \noalign{\smallskip}
            VAE (Recon.) \rule{0pt}{10pt} & 83.6 & 83.3 & 83.2 \\
            \noalign{\smallskip}
            \hline
        \end{tabular}
        \bigskip
        \caption{Effect of logit-normal shift $\mu$ on layout adherence across aspect ratios ($r_{\mathrm{max}}$).}
        \label{tab:mu_ablation_aspect_ratio}
    \end{minipage}
\end{table}

\smallskip
\noindent
\textbf{Results.} We find that the location parameter $\mu$ strongly influences spatial grounding (Tab.~\ref{tab:mu_ablation_resolution}). Fig.~\ref{fig:qualitative_mu_sweep} shows that lower values ($\mu \leq 0.5$) result in perceivable structural misalignments. While increasing $\mu$ to 3.0 yields visually stable layouts, we observe a quantitative performance degradation in $\mathrm{mIoU}_{2:25}$ (Tab.~\ref{tab:mu_ablation_resolution}, Fig.~\ref{fig:layout_adherence_experiment_4_1}). Our analysis (see Appendix~\ref{sec:pixel_precision_examples}) reveals that this drop stems not from structural failure, but from a loss of pixel-level color precision. At this extreme, the model's color assignments frequently fall just outside color-matching threshold $\tau$, despite correct spatial placement. This observation highlights a fundamental trade-off in the $\mu$-shifting schedule: while a high $\mu$ prioritizes global structural grounding, it may limit the model's capacity to resolve high-frequency components. In the context of natural images, this would manifest as a loss of fine-grained textural details. Importantly, moderate shifts ($\mu \in [1.0, 2.5]$) nearly reach the VAE reconstruction upper bound of 83.6. These results suggest that vanilla flow-matching architectures possess the latent capacity to internalize coordinate-based constraints directly from text prompts. Within our controlled synthetic environment, the model achieves near-optimal layout adherence through distribution shifting alone, demonstrating the architectural feasibility of precise spatial grounding without auxiliary modules or inference-time interventions.

\subsection{Resolution \& Aspect Ratio}
\label{sec:results_aspect_ratio_resolution}
To evaluate if this architectural feasibility generalizes beyond fixed square layouts, we extend our approach along two axes: resolution and aspect ratio, which we investigate progressively ($256^2 \rightarrow 512^2 \rightarrow 1024^2$ and $r_{\mathrm{max}} : 1 \rightarrow 2 \rightarrow 3$). For resolution experiments, we re-render our image data at $512^2$ and $1024^2$ while keeping all geometric and textual properties fixed, isolating the effect of resolution from layout complexity. For aspect ratio experiments, we derive resolution buckets $(w, h)$ from a fixed area $A = 256^2$ such that $w \cdot h \approx A$, where $w$ and $h$ are multiples of 16. We include all pairs satisfying $\frac{\max(w, h)}{\min(w, h)} \leq r_{\mathrm{max}}$, ensuring portrait-landscape symmetry. Following prior work \cite{hoogeboom2023simple, esser_stable_diffusion_3, labs2025flux1kontextflowmatching} suggesting that higher resolutions require increased timestep shifting for optimal image quality, we use the optimal $\mu$ from the previous configuration as a lower bound for subsequent searches. For instance, having found $\mu=1$ optimal for $256^2$, we limit our search to $\mu \in [1, 3]$ for the $512^2$ scale. By applying this strategy, we can substantially reduce the search space compared to a full $\mu$-sweep for each setting. Crucially, as bounding box coordinates remain normalized to $[0, 1000]$, the textual prompts are identical across all configurations. This ensures that layout, object count, and color assignments remain constant, allowing for a controlled comparison of spatial grounding performance across varying scales and geometries.

\noindent
\textbf{Resolution.} Table~\ref{tab:mu_ablation_resolution} summarizes the impact of the logit-normal shift $\mu$ on layout adherence as resolution increases. We observe a clear trend: the optimal $\mu$ value is positively correlated with the image resolution. For the $256^2$ base resolution, a moderate shift of $\mu=1.0$ yields the highest performance ($80.8~\mathrm{mIoU}_{2:25}$), nearly saturating the VAE reconstruction bound of $83.6$. However, this configuration fails to generalize to higher resolutions. At $512^2$, $\mu=1.0$ results in a significant drop in layout adherence ($48.1~\mathrm{mIoU}_{2:25}$). Instead, higher resolutions require a more aggressive shift toward earlier timesteps (higher noise levels) to maintain spatial control. Specifically, peak performance shifts to $\mu=2.0$ for $512^2$ and $\mu=2.5$ for $1024^2$. Notably, with the appropriately tuned $\mu$, our model maintains high layout adherence across all scales, reaching $82.7~\mathrm{mIoU}_{2:25}$ at $1024^2$, approaching the VAE's empirical ceiling. This suggests that as the resolution increases, a stronger prior on the early denoising phase is essential for the acquisition of spatial grounding from text prompts alone.

\noindent
\textbf{Aspect Ratio.} We further evaluate the impact of $\mu$ across varying aspect ratios ($r_{\mathrm{max}}$), as shown in Tab.~\ref{tab:mu_ablation_aspect_ratio}. Consistent with the resolution experiments, we find that the base configuration ($\mu=1.0$) is insufficient for handling non-square geometries, with performance dropping from $80.8$ to $67.8~\mathrm{mIoU}_{2:25}$ when moving from $r_{\mathrm{max}}=1$ to $r_{\mathrm{max}}=2$. By increasing $\mu$ to the $2.0\text{--}2.5$ range, the model recovers strong layout adherence even for extreme aspect ratios ($r_{\mathrm{max}}=3$). While there is a slight decrease in absolute $\mathrm{mIoU}_{2:25}$ as the aspect ratio becomes more extreme (likely due to the increased difficulty of placing objects in narrow or highly elongated canvases), the shift in $\mu$ effectively mitigates the drop in spatial control seen with centered timestep distributions. These results demonstrate that the relationship between the timestep distribution $p(t)$ and spatial grounding is not static, but rather highly sensitive to the geometric complexity of the output space. Collectively, these findings indicate that spatial grounding is heavily dependent on the noise levels encountered during training, necessitating a higher noise prior as image resolution and aspect ratios increase.\\

\noindent
We further find that layout adherence is strongly dependent on object size (tiny vs. large) but largely density-independent, remaining stable for $n \in [3,10]$ objects per image. For tiny objects ($<16^2$ px), median $\mathrm{mIoU}_{2:25}$ is approximately 0.41, versus 0.85-0.95 for large ($64-128^2$ px) and huge ($>128^2$ px) objects. However, a similar degradation appears in the VAE reconstruction ceiling itself, suggesting this size-dependence is a property of the underlying latent space rather than a model-specific limitation. Full results are in Appendix~\ref{sec:results_scale_density}.

\subsection{Architectural Generalization}
To assess if the relationship between spatial grounding and SNR distribution shift (parameterized by $\mu$) generalizes across different architectural configurations and text-conditioning strategies, we extend our evaluation to three larger-scale models: FLUX.2-9B-base \cite{flux-2-2025} and Qwen-Image (20B)~\cite{2025qwenimage}, which both utilize a pre-aligned Vision-Language Model (VLM) encoder, and FLUX.1-dev (12B)~\cite{labs2025flux1kontextflowmatching, flux2024}, which employs a dual-encoder setup consisting of CLIP-L~\cite{Radford2021CLIP} and T5-XXL~\cite{Raffel_T5}. Since these models utilize different VAE architectures with varying reconstruction fidelities, we measure performance relative to their respective VAE upper bounds. We use $\mu=1.0$, the optimal shift identified in Section~\ref{sec:results_aspect_ratio_resolution} for square images at $256^2$ resolution, as a common starting point across architectures and compare it against a centered SNR distribution ($\mu=0.0$) as a baseline.

\begin{table}[t]
    \centering
    \begin{tabular}{lllcccc}
        \noalign{\smallskip}
        & & & & \multicolumn{3}{c}{$\text{mIoU}_{2:25}$} \\ \cmidrule{5-7}
        \noalign{\smallskip}
        Model & Params & Text Encoder & $\mu$ & Model $\uparrow$ & VAE $\uparrow$ & $\Delta (Gap) \downarrow$\\
        \midrule
        \multirow{2}{*}{FLUX.2-4B-base} & \multirow{2}{*}{4B} & \multirow{2}{*}{Qwen3-VL-4B} & $0.0$ & 25.9 & 83.6 & 57.7 \\
        & & & $1.0$ & 80.8 & 83.6 & 2.8 \\
        \midrule
        \multirow{2}{*}{FLUX.2-9B-base}  & \multirow{2}{*}{9B}  & \multirow{2}{*}{Qwen3-VL-8B}    & $0.0$ & 32.6 & 83.6  & 51.0 \\
          & &  & $1.0$ & 81.5 & 83.6 & 2.1 \\
        \midrule
        \multirow{2}{*}{FLUX.1-dev} & \multirow{2}{*}{12B} & \multirow{2}{*}{CLIP + T5} & $1.0$ & 7.5 & 88.6 & 81.1 \\
         & & & $3.0$ & 78.7 & 88.6 & 9.9 \\
        \midrule
        \multirow{2}{*}{Qwen-Image} & \multirow{2}{*}{20B} & \multirow{2}{*}{Qwen2.5-VL-7B} & $0.0$ & 48.1 & 92.5 & 44.4 \\
         &  &  & $1.0$ & 80.8 & 92.5 & 11.7 \\
        \midrule
        \noalign{\medskip}
    \end{tabular}
    \caption{\textbf{Architectural Generalization.} Strong layout adherence can be achieved across scales, architectures, and text-conditioning strategies through SNR shifting alone. $\mu=1.0$ is used as a common starting point across all models, compared against a centered baseline ($\mu=0.0$). FLUX.1-dev instead requires $\mu=3.0$ to reach comparable spatial grounding. "Gap" ($\Delta$) denotes the difference between VAE reconstruction and generated performance.}
    \label{tab:layout_adherence_architectural_generalization}
\end{table}

\noindent
\textbf{Generalization Across Architectures.} The results in Tab.~\ref{tab:layout_adherence_architectural_generalization} demonstrate that shifting the timestep sampling distribution is a generalizable mechanism to unlock spatial grounding in flow-matching models using text prompts only. This effect remains consistent regardless of the specific transformer backbone or text-conditioning strategy. While our primary experiments use the FLUX.2-4B-base architecture, the ability to recover layout adherence in 9B, 12B, and 20B models by tuning the SNR shift suggests a fundamental property of how flow-matching models interpret spatial instructions across scales.

\noindent
\textbf{Impact of Text Conditioning.} 
We observe a clear correlation between the choice of text encoder and the model's sensitivity to the shifting parameter $\mu$. At $\mu = 1.0$, models using Vision-Language Model (VLM) encoders, such as Qwen-Image and FLUX.2, demonstrate high layout adherence ($\geq80.8~\mathrm{mIoU}_{2:25}$). In contrast, FLUX.1-dev (using T5 and CLIP) requires a significantly stronger shift ($\mu=3.0$) to achieve comparable grounding performance. This suggests that VLM-based conditioning provides a more integrated spatial-semantic representation that is inherently more "accessible" to the denoising process than the separately trained embeddings provided by T5 and CLIP encoders.

\noindent
\textbf{Generative Gap and VAE Learnability.} While all four models achieve high $\mathrm{mIoU}_{2:25}$ scores, their "generative gaps" ($\Delta$) reveal a fundamental trade-off between VAE reconstruction fidelity and latent-space learnability. The FLUX.2 VAE is specifically optimized for representation learnability~\cite{bfl2025representation}, which is reflected in the remarkably small gaps of 2.8 (FLUX.2-4B-base) and 2.1 (FLUX.2-9B-base), indicating that the FLUX.2 models can almost fully exploit the empirical VAE reconstruction limit. However, this optimization for learnability entails a trade-off in reconstruction precision (e.g., color fidelity), leading to a lower VAE upper bound compared to Qwen-Image or FLUX.1-dev. Conversely, while the latter models possess higher reconstruction ceilings, their latent spaces appear significantly harder to navigate for precise spatial grounding, as evidenced by their larger generative gaps. This suggests that recent advancements in improving VAE learnability~\cite{zheng2026diffusion,bfl2025representation} not only benefit image quality but also play a critical role in a model's ability to adhere to complex layout instructions from text prompts.

\section{Real-World Images}
\label{sec:real_world_results}
In the previous section, we established that flow-matching models can acquire strong layout adherence through text prompts alone by optimizing the SNR distribution. While our synthetic benchmark allowed for a controlled mechanistic study, we now investigate whether these findings transfer to the semantically rich domain of real-world images.

\noindent\textbf{Dataset and Evaluation.} Following standardized protocols~\cite{zhang2025creatilayout, Yang_2023_ICCV, Li_GLIGEN_2023_CVPR}, we evaluate on MS-COCO 2017~\cite{lin2014microsoft}, excluding images with crowds, dense clusters, or objects smaller than 2\% of the canvas. This yields 61,986 training and 2,565 validation images. We assess layout adherence using a frozen YOLO-v11-x~\cite{khanam2024yolov11} object detector to report Average Precision ($\mathrm{AP}$, $\mathrm{AP}^{50}$) and Average Recall ($\mathrm{AR}$), while reporting the detector's performance on ground-truth images as an empirical ceiling. Additionally, we assess global image quality and semantic alignment by computing CLIP score~\cite{hessel2021clipscore}, FID~\cite{heusel2017gans}, and Inception Score (IS)~\cite{salimans2016improved}.

\begin{table}[t]
    \centering
    \begin{tabular}{lcccccc}
        \noalign{\smallskip}
        & \multicolumn{3}{c}{Layout Adherence} & \multicolumn{3}{c}{Image Quality}\\
        \noalign{\smallskip}\cmidrule(r){2-4}\cmidrule(l){5-7}\noalign{\smallskip}
        Method & $\mathrm{AP}\uparrow$ & $\mathrm{AP}^{50}\uparrow$ & $\mathrm{AR}\uparrow$ & $\mathrm{CLIP}\uparrow$ & $\mathrm{FID}\downarrow$ & $\mathrm{IS}\uparrow$\\
        \noalign{\smallskip}\hline\noalign{\smallskip}
        $p_{ln}(t; \mu=3.0, \cdot)$ & \textbf{57.3} & \textbf{81.8} & \textbf{65.2} & \textbf{28.01} & \underline{32.38} & \textbf{30.32} \\
        $p_{ln}(t; \mu=2.5, \cdot)$ & \underline{52.8} & \underline{78.9} & \underline{60.5} & \underline{27.92} & 32.44 & 29.62 \\
        $p_{ln}(t; \mu=2.0, \cdot)$ & 32.0 & 61.6 & 40.9 & 27.85 & 32.45 & 29.38 \\
        $p_{ln}(t; \mu=1.5, \cdot)$ & 14.9 & 36.3 & 23.1 & 27.65 & \textbf{32.00} & \underline{29.82} \\
        $p_{ln}(t; \mu=1.0, \cdot)$ & 1.4 & 4.8 & 5.4 & 27.04 & 33.11 & 29.27 \\
        $p_{ln}(t; \mu=0.5, \cdot)$ & 1.0 & 3.1 & 4.2 & 26.97 & 34.34 & 27.84 \\
        $p_{ln}(t; \mu=0.0, \cdot)$ & 1.0 & 2.8 & 3.7 & 26.93 & 37.75 & 27.62 \\
        \hline
        \noalign{\smallskip}
        COCO val (GT)                               & 61.6 & 77.2  & 74.9 & 30.47 & 0.00  & 29.06 \\
        \hline
        \noalign{\medskip}
    \end{tabular}
    \caption{\textbf{Effect of shift $\mu$ on layout adherence and image quality of real-world images.} To validate generalization beyond our synthetic benchmark, FLUX.2-9B-base is fine-tuned for 5 epochs on MS-COCO 2017~\cite{lin2014microsoft} at $512^2$ resolution and aspect ratios up to 3:1 ($r_{\mathrm{max}} {=} 3$).}
    \label{tab:mu_ablation_real_world}
\end{table}

\begin{table}[H]
    \centering
    \begin{tabular}{lcccccc}
        \noalign{\smallskip}
        & \multicolumn{3}{c}{Layout Adherence} & \multicolumn{3}{c}{Image Quality}\\
        \noalign{\smallskip}\cmidrule(r){2-4}\cmidrule(l){5-7}\noalign{\smallskip}
        Method & $\mathrm{AP}\uparrow$ & $\mathrm{AP}^{50}\uparrow$ & $\mathrm{AR}\uparrow$ & $\mathrm{CLIP}\uparrow$ & $\mathrm{FID}\downarrow$ & $\mathrm{IS}\uparrow$\\
        \noalign{\smallskip}\hline\noalign{\smallskip}
        GLIGEN~\cite{Li_GLIGEN_2023_CVPR}           & 42.1 & 66.8 & 37.2 & 30.25 & 30.68 & 26.76 \\
        MIGC~\cite{Zhou_2024_CVPR}                  & 39.1 & 66.0 & 36.8 & 28.60 & 31.43 & 26.14 \\
        InstanceDiff~\cite{Wang_2024_CVPR}          & 51.9 & 71.9 & 43.7 & 30.04 & \underline{29.94} & 27.23 \\
        Ranni~\cite{Feng_2024_CVPR}                 & 14.7 & 22.3 & 17.0 & 30.02 & 31.94 & 24.44 \\
        CreatiLayout~\cite{zhang2025creatilayout}   & 47.8 & 69.8 & 42.1 & \textbf{33.32} & \textbf{29.27} & \underline{30.26} \\
        Hybrid Layout~\cite{wu2025hybrid}   & 55.2 & 79.1 & 51.8 & \underline{31.20} & -- & 26.72 \\
        \noalign{\vspace{0.05cm}}
        \multicolumn{7}{l}{\textit{FLUX.2-9B-base}} \\
        \noalign{\vspace{-0.06cm}}
        \enspace$\mu{=}3.0,\; \textrm{epochs}{=}5$   & \underline{57.3} & \underline{81.8} & \underline{65.2} & 28.01 & 32.38 & \textbf{30.32} \\
        \enspace$\mu{=}3.0,\; \textrm{epochs}{=}10$  & \textbf{61.8}    & \textbf{84.7}    & \textbf{68.8}    & 28.03 & 31.43 & 30.06 \\
        \noalign{\smallskip}
        \hline
        \noalign{\smallskip}
        COCO val (GT)                               & 61.6 & 77.2  & 74.9 & 30.47 & 0.00  & 29.06 \\
        \noalign{\smallskip}
        \hline
        \noalign{\medskip}
    \end{tabular}
    \caption{\textbf{Quantitative comparison against specialized layout-to-image models.} Evaluated on MS-COCO 2017, our fine-tuned FLUX.2-9B-base achieves state-of-the-art layout adherence and competitive image quality, without requiring architectural modifications.}
    \label{tab:layout_adherence_real_world_imagery}
\end{table}

\noindent\textbf{Implementation Details.} We select FLUX.2-9B-base for our real-world experiments, as it achieved the strongest layout adherence among the architectures evaluated in Tab.~\ref{tab:layout_adherence_architectural_generalization} ($81.5$ $\mathrm{mIoU}_{2:25}$ at $\mu=1.0$). We fine-tune this model for 5 epochs at $512^2$ resolution, supporting aspect ratios up to 3:1 ($r_{\mathrm{max}} = 3$). As done for our synthetic benchmark (Tab.~\ref{tab:mu_ablation_resolution}, \ref{tab:mu_ablation_aspect_ratio}), we investigate $\mu \in \left\{0.0, 0.5, 1.0, 1.5, 2.0, 2.5, 3.0 \right\}$. To preserve the base model's generative priors amidst complex real-world textures, we use a conservative learning rate of $1\times 10^{-6}$, which we found yields more stable convergence than the higher rates used for geometric primitives. For real-world images, we adopt the standard FLUX.2 sampling configuration (50 steps, 4.0 guidance). While more efficient settings sufficed in our synthetic study, these default parameters are better suited to resolving complex real-world textures. Bounding boxes are normalized to $[0, 1000]$, sequenced by appearance in the COCO annotations, and formatted as: $[x_{min}, y_{min}, x_{max}, y_{max}, \text{category name}]$.

\begin{figure}[t]
    \centerline{\includegraphics[width=\linewidth]{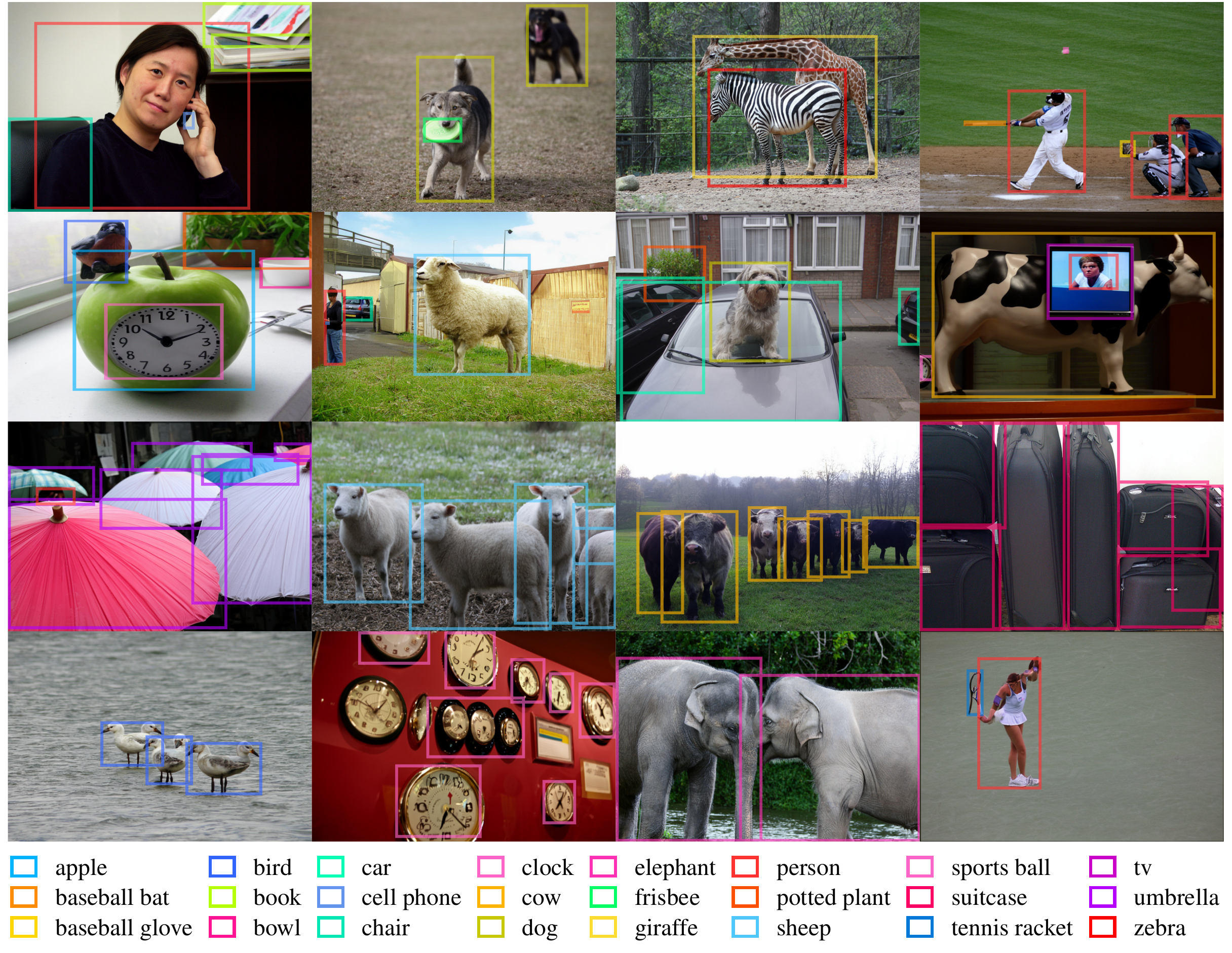}}
    \caption{\textbf{Qualitative results on COCO val.} Bounding boxes indicate ground-truth layout conditions. Our model faithfully generates objects at specified locations across diverse scenes, compositions, and densities (rows 1--3). Row 4 shows representative failure cases.}
    \label{fig:qualitative_real_world_results}
\end{figure}

\noindent\textbf{Results.} Table~\ref{tab:mu_ablation_real_world} demonstrates that spatial grounding via SNR distribution shifting ($\mu=3.0$) robustly transfers from our controlled synthetic environment to complex real-world imagery. Despite using only text prompts and a lightweight 5-epoch fine-tuning on MS-COCO, our model achieves state-of-the-art layout adherence compared to specialized layout-to-image models (Table~\ref{tab:layout_adherence_real_world_imagery}). Specifically, our Average Recall ($\mathrm{AR}$) of $65.2$ significantly outperforms established models like GLIGEN ($37.2$), InstanceDiff ($43.7$), or Hybrid Layout ($51.8$), indicating that our model is remarkably effective at populating the canvas with all requested objects. Furthermore, our $\mathrm{AP}^{50}$ of $81.8$ exceeds the YOLO-v11-x detector's empirical ceiling on ground-truth COCO images ($77.2$). Since ground-truth represents a theoretical upper bound, we hypothesize this reflects known limitations in the COCO dataset itself (e.g., bounding box noise, motion blur, and greater occlusion and background clutter than in our generated images) that lower the detector's measured ceiling on real-world images. Overall, this shows that our model's spatial precision is effectively saturating the empirical ceiling. While FID ($32.38$) and CLIP scores ($28.01$) trail CreatiLayout~\cite{zhang2025creatilayout}, which utilizes a more complex multi-stream architecture, our model achieves the highest Inception Score ($30.32$) among all compared methods, including CreatiLayout ($30.26$) and the ground-truth COCO images themselves ($29.06$), indicating that the rich generative priors of the FLUX.2-9B-base model are preserved during fine-tuning. To assess whether extended training yields further gains, we fine-tune FLUX.2-9B-base at $\mu=3.0$ for 10 epochs while evaluating after each epoch, which further improves layout adherence across all metrics ($\mathrm{AP}$: $61.8$, $\mathrm{AP}^{50}$: $84.7$, $\mathrm{AR}$: $68.8$; Tab.~\ref{tab:layout_adherence_real_world_imagery}). The complete epoch-ablation is reported in Appendix~\ref{sec:epoch_ablation}. Qualitative results (Fig.~\ref{fig:qualitative_real_world_results}) of the FLUX.2-9B-base model after 10 epochs of fine-tuning show faithful spatial generation across diverse scenes, challenging compositions, and densities (rows 1--3). Observed failure modes (row 4) include compositional artifacts (e.g., merged or duplicated anatomical features) and high-frequency imprecision (e.g., clock face details). Preliminary experiments applying layout-conditioned denoising only during the first 40--50\% of steps improved fine-grained details with only a marginal $\mathrm{AP}$ drop ($1.5$--$5.1$ points). While some compositional artifacts persist, also occasionally seen in the pre-trained FLUX.2-9B-base model prior to any fine-tuning (Appendix~\ref{sec:compositional_failures}), unifying these dynamics into a single, potentially scaled, foundation model remains future work. Ultimately, these findings validate the insights from our synthetic benchmark: the lack of native spatial control in foundation flow-matching models is not an architectural limitation. By shifting the SNR distribution to prioritize global structural dependencies, a standard flow-matching transformer can parse spatial coordinates directly from the text stream and compete with specialized architectures.

%% file: chapters/conclusion.tex
\section{Conclusion}
\label{sec:conclusion}
In this work, we investigated the latent spatial grounding capabilities of flow-matching models, addressing the discrepancy between the rich spatial embeddings of modern vision-language encoders and the poor native layout adherence of the generative models that leverage them. Unlike the prevailing paradigm of introducing auxiliary structural adapters or complex inference-time optimizations, we show that precise spatial control can be "unlocked" by shifting the timestep sampling distribution. Our systematic evaluation using a novel synthetic benchmark of geometric primitives allowed us to isolate the mechanics of spatial grounding from textural complexities. We identified the high-noise regime as a "critical learning phase", where the model is forced to internalize geometric constraints to establish global structure. By biasing the fine-tuning process toward these regimes, we enabled models to interpret 2D coordinates directly from text prompts with strong spatial accuracy. Furthermore, we demonstrated that these insights generalize to real-world imagery via the COCO dataset, achieving state-of-the-art layout adherence without any architectural modifications. The implications of our findings are twofold: first, they suggest that the perceived lack of spatial control in current models is not a limitation of architectural capacity. Second, our approach offers a highly efficient, architecture-agnostic path for enhancing spatial controllability, providing a recipe for integrating native layout adherence into future foundation models.\\

\noindent
\textbf{Limitations and future work.} While our approach provides a lightweight and architecture-agnostic path to spatial grounding, it reveals a trade-off: a strong bias toward high-noise regimes during fine-tuning, while essential for layout adherence, can compromise high-frequency textural synthesis. This suggests that the global structural alignment facilitated by high-noise training may compete with local generative refinement required for high-fidelity synthesis. Future research should investigate multi-stage or scheduled sampling distributions that transition from a high-noise bias (to lock in grounding) to a more uniform distribution (to recover textural quality). Furthermore, exploring Direct Preference Optimization (DPO)~\cite{liu2026improving} or other reinforcement learning techniques~\cite{liu2026flow} presents a promising avenue for aligning model outputs with both layout constraints and human-perceived image quality.

%% file: chapters/appendix.tex
\section{Loss of Pixel-Level Color Precision at High $\mu$}
\label{sec:pixel_precision_examples}

\begin{figure}[h]
    \centerline{\includegraphics[width=\linewidth]{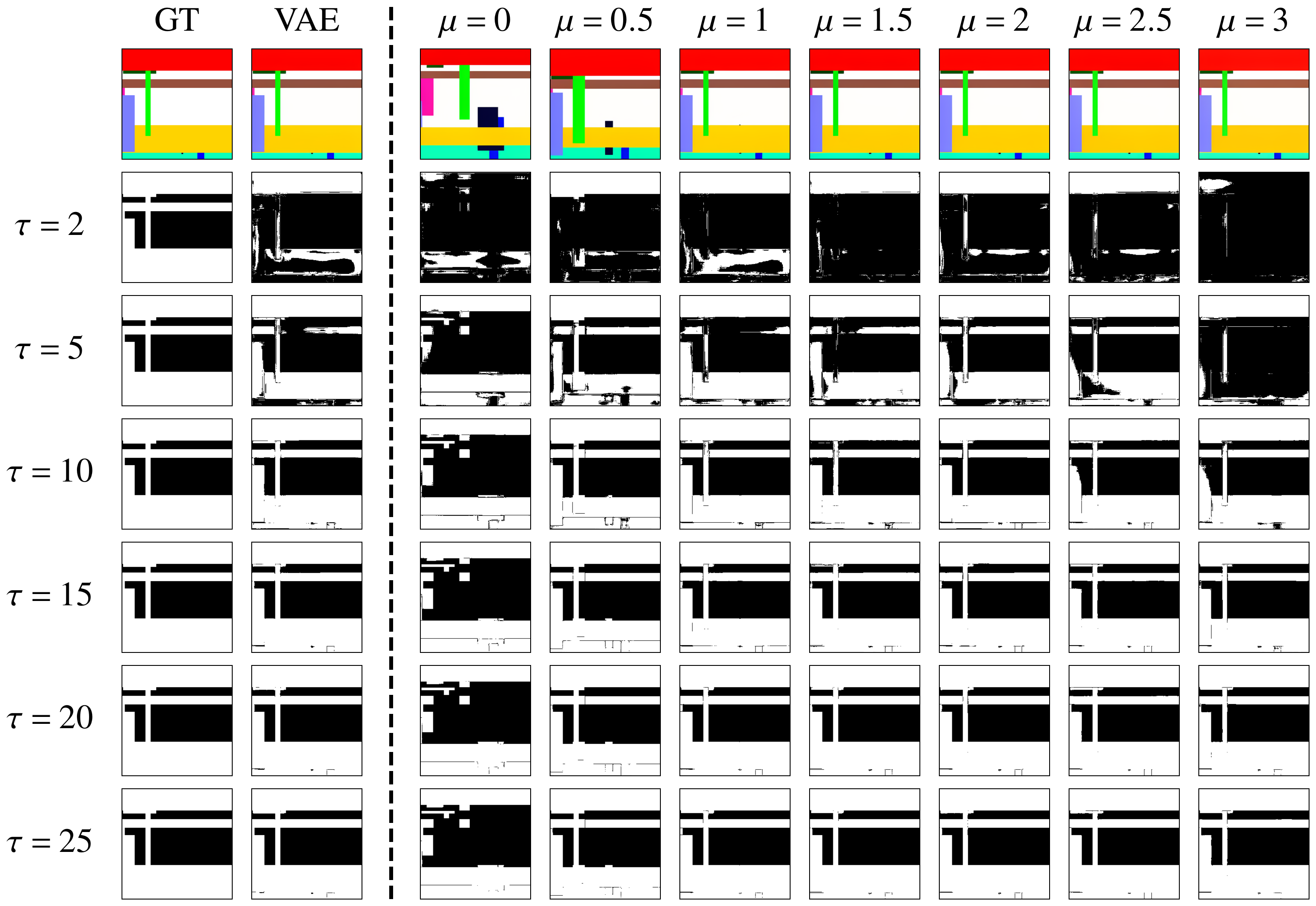}}
    \caption{\textbf{Loss of pixel-level color precision at $\mu=3.0$ (Example 1/3).} Aggregated per-color binary mask intersections across thresholds $\tau \in \{2,5,10,15,20,25\}$ (rows) and GT/VAE/generated images at varying $\mu$ (columns). Intersection masks at $\mu=3.0$ fade sharply for low $\tau$ ($\leq 5$), indicating loss of color precision rather than spatial misplacement.}
    \label{fig:pixel_precision_example_1}
\end{figure}

\begin{figure}[H]
    \centerline{\includegraphics[width=\linewidth]{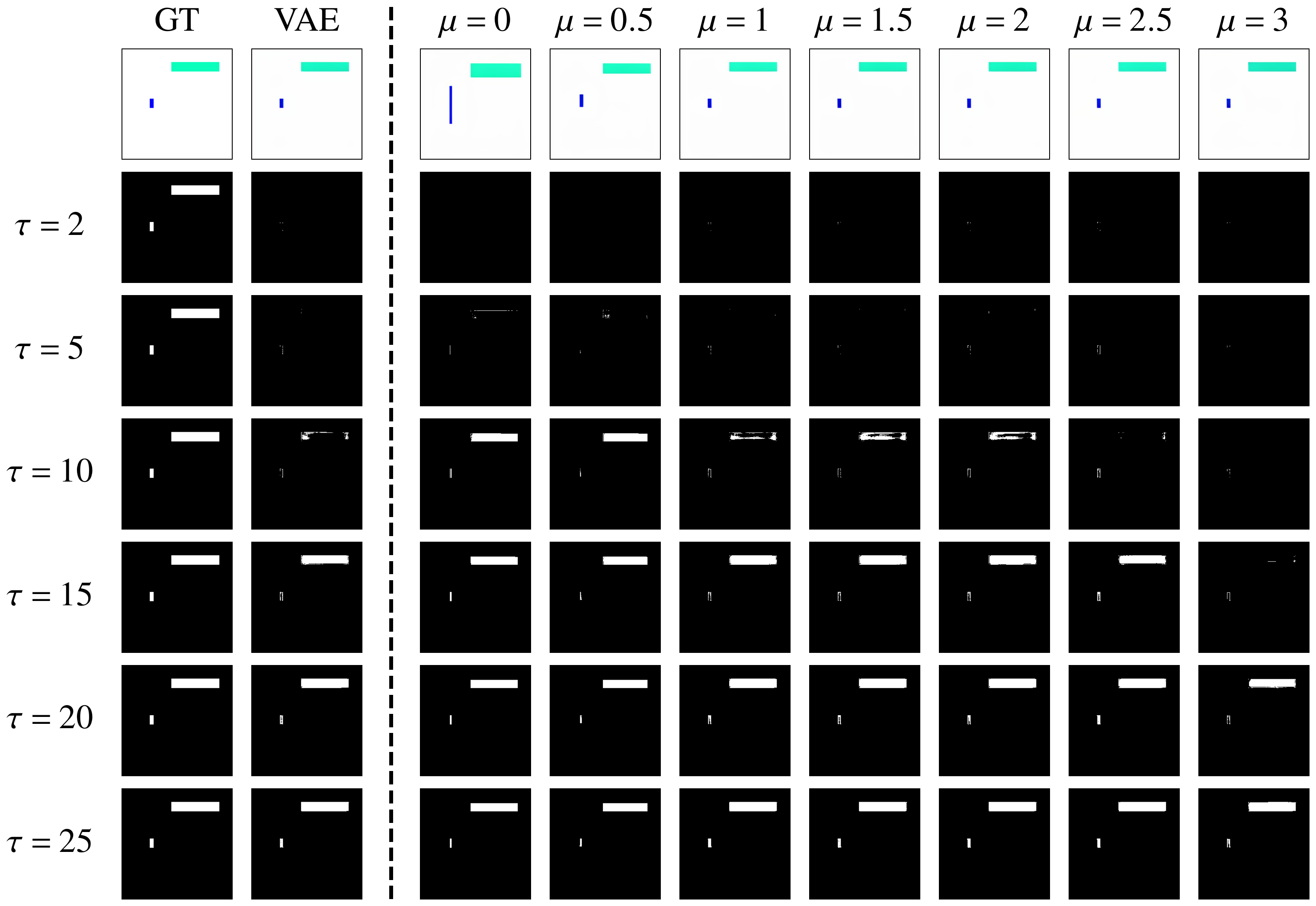}}
    \caption{\textbf{Loss of pixel-level color precision at $\mu=3.0$ for $\tau \leq 15$ in a sparse scene (2/3).}}
    \label{fig:pixel_precision_example_2}
\end{figure}

\begin{figure}[H]
    \centerline{\includegraphics[width=\linewidth]{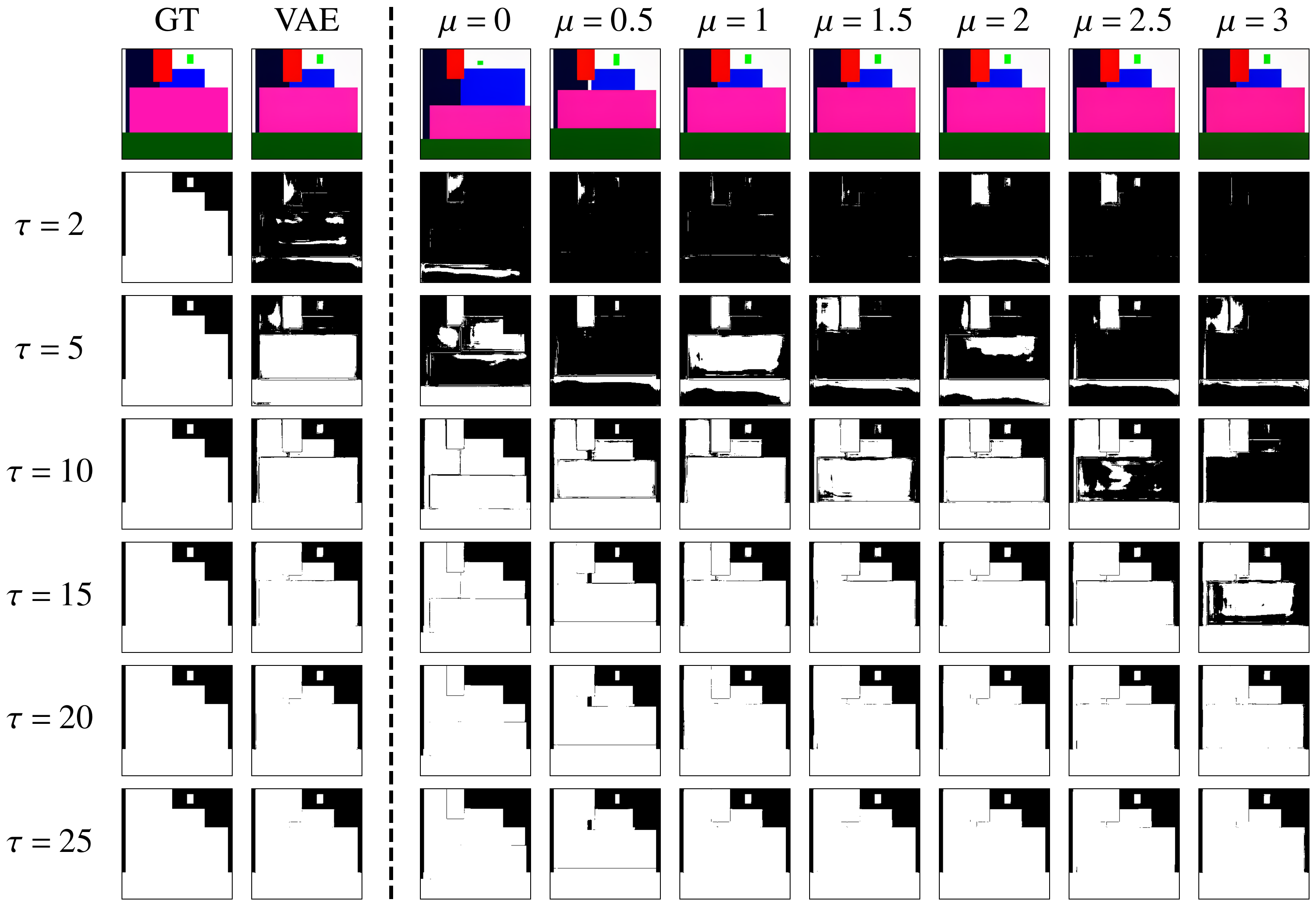}}
    \caption{\textbf{Loss of pixel-level color precision at $\mu=3.0$ for $\tau \leq 15$ in a dense scene (3/3).}}
    \label{fig:pixel_precision_example_3}
\end{figure}

To further understand the quantitative drop in $\mathrm{mIoU}_{2:25}$ observed with shifting parameter $\mu=3.0$ for square images at $256\times256$ resolution on our synthetic benchmark (Sec.~\ref{sec:snr_dist}), we conducted a manual visual analysis of generated images under this setting. We observed that the drop does not come from structural failure, but from a loss of pixel-level color precision. The model's color assignments frequently fall just outside the color-matching threshold $\tau$, despite correct spatial placement. While examples of correct spatial placement at $\mu=3.0$ for square images at $256\times256$ resolution were already shown in Fig.~\ref{fig:qualitative_mu_sweep}, we provide additional examples in Fig.~\ref{fig:pixel_precision_example_1}--\ref{fig:pixel_precision_example_3} that visualize the aggregated (union of) per-color binary intersection masks $\bigcup_{c \in \mathcal{C}} \left[ M_c(I_{\mathrm{gt}}; \tau) \cap M_c(I_{\mathrm{gen}}; \tau) \right]$, which together visualize the intersection component of the per-color $\mathrm{IoU}_{c}(I_{\mathrm{gt}}, I_{\mathrm{gen}}; \tau)$ (Eq.~\ref{eq:iou_c}) that is averaged into $\mathrm{mIoU}_\tau$ (Eq.~\ref{eq:miou_tau}, Sec.~\ref{subsec:layout_adherence_metric}). Aggregating across colors thus reveals which image regions are correctly matched at a given threshold $\tau$. Each figure shows a grid of these aggregated intersection masks, with rows corresponding to color-matching threshold values $\tau \in \{2, 5, 10, 15, 20, 25\}$ and columns corresponding to the ground-truth image, its VAE reconstruction, and generated images (using the fine-tuned FLUX.2-4B-base model) at different values of $\mu$. In these figures, it can be seen that generated images for $\mu \geq 1$ are spatially and compositionally nearly identical. However, for the same $\tau$, intersection coverage differs sharply across $\mu$. For $\mu=3.0$, color assignments frequently fall just outside the threshold (especially for low $\tau$), while for $\mu \leq 2.5$ they achieve higher color precision.

\section{Impact of Object Scale \& Density on Layout Adherence}
\label{sec:results_scale_density}
To assess how layout adherence varies with scene complexity and object scale, we evaluate the impact of object count and object size on the resulting spatial accuracy. For the object count analysis, we evaluate each value of $n \in \{1, \ldots, 10\}$ individually, corresponding to the full range of rectangles per image in our benchmark. For the size analysis, we categorize individual rectangles into five categories: \textit{Tiny} ($< 16^{2}$ px), \textit{Small} ($16^{2} \text{--} 32^{2}$ px), \textit{Medium} ($32^{2} \text{--} 64^{2}$ px), \textit{Large} ($64^{2} \text{--} 128^{2}$ px), and \textit{Huge} ($>128^{2}$ px).

\begin{figure}[t]
    \centerline{\includegraphics[width=\linewidth]{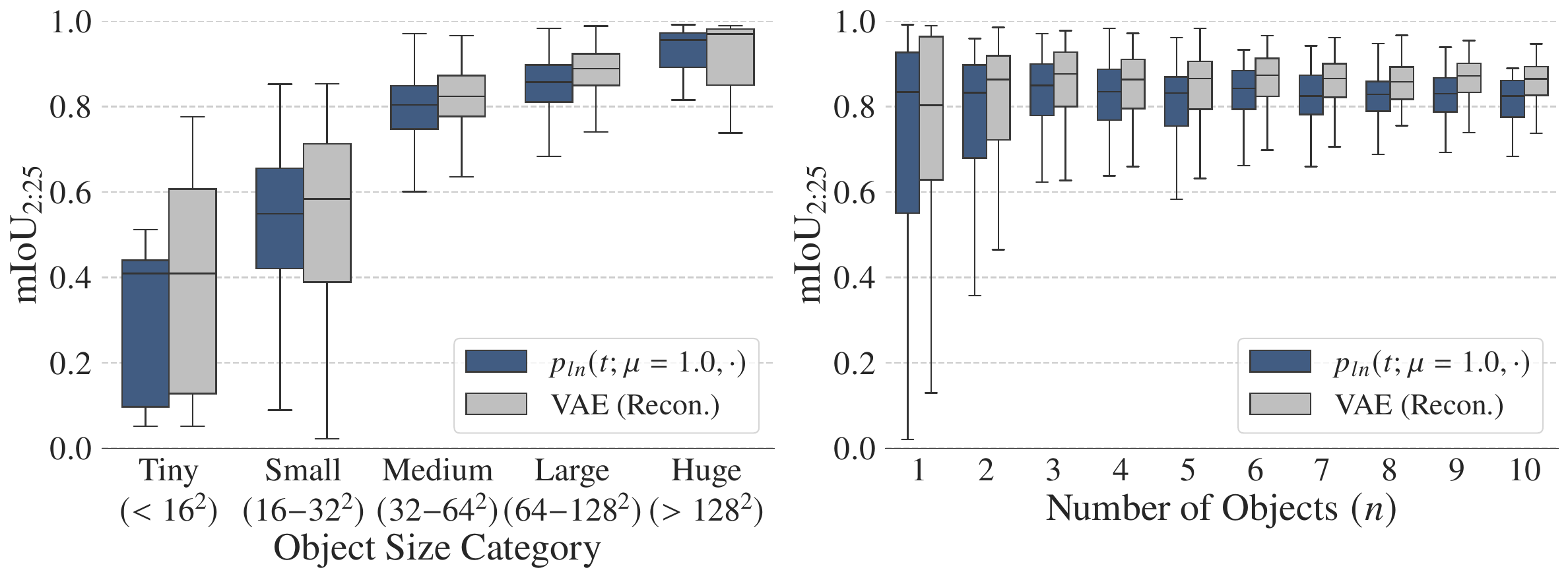}}
    \caption{\textbf{Scale and density-dependent layout adherence.} The generative model (blue) saturates the VAE's capacity (gray). While localization is harder for smaller objects, performance is stable across scene densities, with near-constant median $\mathrm{mIoU}_{2:25}$ for $n \in [3,10]$.}
    \label{fig:object_size_scale_results}
\end{figure}

\noindent
\textbf{Object Scale.} The relationship between object scale and layout adherence is illustrated in Fig.~\ref{fig:object_size_scale_results} (left). We observe a positive correlation between object size and spatial accuracy. As rectangle area increases, the model's $\mathrm{mIoU}_{2:25}$ consistently improves. For \textit{Large} and \textit{Huge} objects, the model achieves near-optimal spatial fidelity, with median $\mathrm{mIoU}_{2:25}$ ranging from 0.85 to 0.95. While performance remains robust for \textit{Medium} objects, we note an increase in variance, with the lower bound of the distribution dropping to 0.60. The model's primary failure modes occur at smaller scales, where \textit{Tiny} objects drop to a median $\mathrm{mIoU}_{2:25}$ of approximately 0.41. Crucially, as shown in Fig.~\ref{fig:object_size_scale_results}, our model's median performance closely tracks the VAE reconstruction baseline across all size categories, indicating that the model effectively approaches the average spatial capacity of the underlying latent representation. However, we note a slight divergence in the upper quartiles (75th percentile) for \textit{Tiny} and \textit{Small} objects, where the VAE baseline maintains a higher performance ceiling. This suggests that while our model accurately predicts the expected location of \textit{Small} objects, it does not yet fully capture the extreme precision occasionally preserved by the VAE reconstruction.

\noindent
\textbf{Object Density.} We further analyze the impact of object density on layout adherence in Fig.~\ref{fig:object_size_scale_results} (right). Counter-intuitively, the model exhibits its highest variance when few rectangles are present ($n=1, 2$), where the $\mathrm{mIoU}_{2:25}$ distribution is broadest. Our investigation reveals that this is primarily driven by the object size sensitivity discussed previously. Performance in single-object images is highly dependent on whether that specific object is \textit{Tiny} or \textit{Huge}. As the number of objects increases, the statistics stabilize, with the median $\mathrm{mIoU}_{2:25}$ remaining remarkably constant up to $n=10$. This suggests that the model is robust to scene complexity and effectively maintains layout adherence in crowded environments, consistently approaching the VAE reconstruction median regardless of object count.

\section{Extended Fine-Tuning on Real-World Images}
\label{sec:epoch_ablation}
To assess whether layout adherence continues to improve beyond the 5-epoch budget used in our $\mu$-ablation (Tab.~\ref{tab:mu_ablation_real_world}), we fine-tune FLUX.2-9B-base at $\mu=3.0$ for 10 epochs, evaluating after each. Full epoch-by-epoch results are shown in Tab.~\ref{tab:epoch_ablation_real_world}. The 10-epoch endpoint is also reported in Tab.~\ref{tab:layout_adherence_real_world_imagery}. Note that the 5-epoch checkpoint comes from this separate extended run and therefore slightly differs from the $\mu$-ablation's 5-epoch result (Tab.~\ref{tab:mu_ablation_real_world}, Tab.~\ref{tab:layout_adherence_real_world_imagery}) due to training variance. Average Precision (AP) and Average Recall (AR) improve continuously across all 10 epochs, with AP$^{50}$ peaking at epoch 9, suggesting diminishing returns beyond this point. AP and AP$^{50}$ exceed ground-truth COCO values after 10 epochs (see Sec.~\ref{sec:real_world_results} for possible explanations), while AR approaches ground-truth. Image quality metrics vary more: CLIP peaks at epoch 7, remaining fairly stable between epochs 3--8 before declining slightly. The Inception Score fluctuates more but is generally highest from epoch 5 onward. FID, while also fluctuating, shows a general improving trend with extended training.

\begin{table}[t]
    \centering
    \begin{tabular}{ccccccc}
        \noalign{\smallskip}
        & \multicolumn{3}{c}{Layout Adherence} & \multicolumn{3}{c}{Image Quality}\\
        \noalign{\smallskip}\cmidrule(r){2-4}\cmidrule(l){5-7}\noalign{\smallskip}
        Epoch & $\mathrm{AP}\uparrow$ & $\mathrm{AP}^{50}\uparrow$ & $\mathrm{AR}\uparrow$ & $\mathrm{CLIP}\uparrow$ & $\mathrm{FID}\downarrow$ & $\mathrm{IS}\uparrow$\\
        \noalign{\smallskip}\hline\noalign{\smallskip}
        $0$  & 0.3  & 0.9  & 1.6  & 22.90 & 78.99 & 18.29 \\
        $1$  & 17.5 & 40.1 & 26.0 & 27.83 & 33.78 & 29.46 \\
        $2$  & 36.4 & 66.1 & 45.1 & 27.91 & 32.89 & 29.57 \\
        $3$  & 45.5 & 73.7 & 53.7 & \underline{28.06} & 31.55 & 29.19 \\
        $4$  & 51.1 & 78.3 & 59.1 & \underline{28.06} & 31.88 & 29.69 \\
        $5$  & 54.6 & 80.5 & 62.5 & \underline{28.06} & 32.21 & 30.06 \\
        $6$  & 57.0 & 82.5 & 64.7 & \underline{28.06} & 31.47 & \textbf{30.47} \\
        $7$  & 58.7 & 83.8 & 66.2 & \textbf{28.08} & 31.34 & \underline{30.20} \\
        $8$  & 59.9 & 84.5 & 67.2 & \underline{28.06} & \underline{31.11} & 29.78 \\
        $9$  & \underline{60.7} & \textbf{85.0} & \underline{68.1} & 28.02 & \textbf{31.06} & 29.36 \\
        $10$ & \textbf{61.8} & \underline{84.7} & \textbf{68.8} & 28.03 & 31.43 & 30.06 \\
        \noalign{\smallskip}
        \hline
        \noalign{\smallskip}
        COCO val (GT)                               & 61.6 & 77.2  & 74.9 & 30.47 & 0.00  & 29.06 \\
        \hline
        \noalign{\medskip}
    \end{tabular}
    \caption{\textbf{Epoch-ablation of layout adherence and image quality on real-world images.} FLUX.2-9B-base is fine-tuned on MS-COCO 2017~\cite{lin2014microsoft} at $\mu=3.0$ and evaluated after each of 10 epochs. Results at $E=5$ slightly differ from Tab.~\ref{tab:mu_ablation_real_world} and Tab.~\ref{tab:layout_adherence_real_world_imagery} due to training variance.}
    \label{tab:epoch_ablation_real_world}
\end{table}

\section{Compositional Failures in Pre-Trained Model}
\label{sec:compositional_failures}
\begin{figure}[t]
    \centerline{\includegraphics[width=\linewidth]{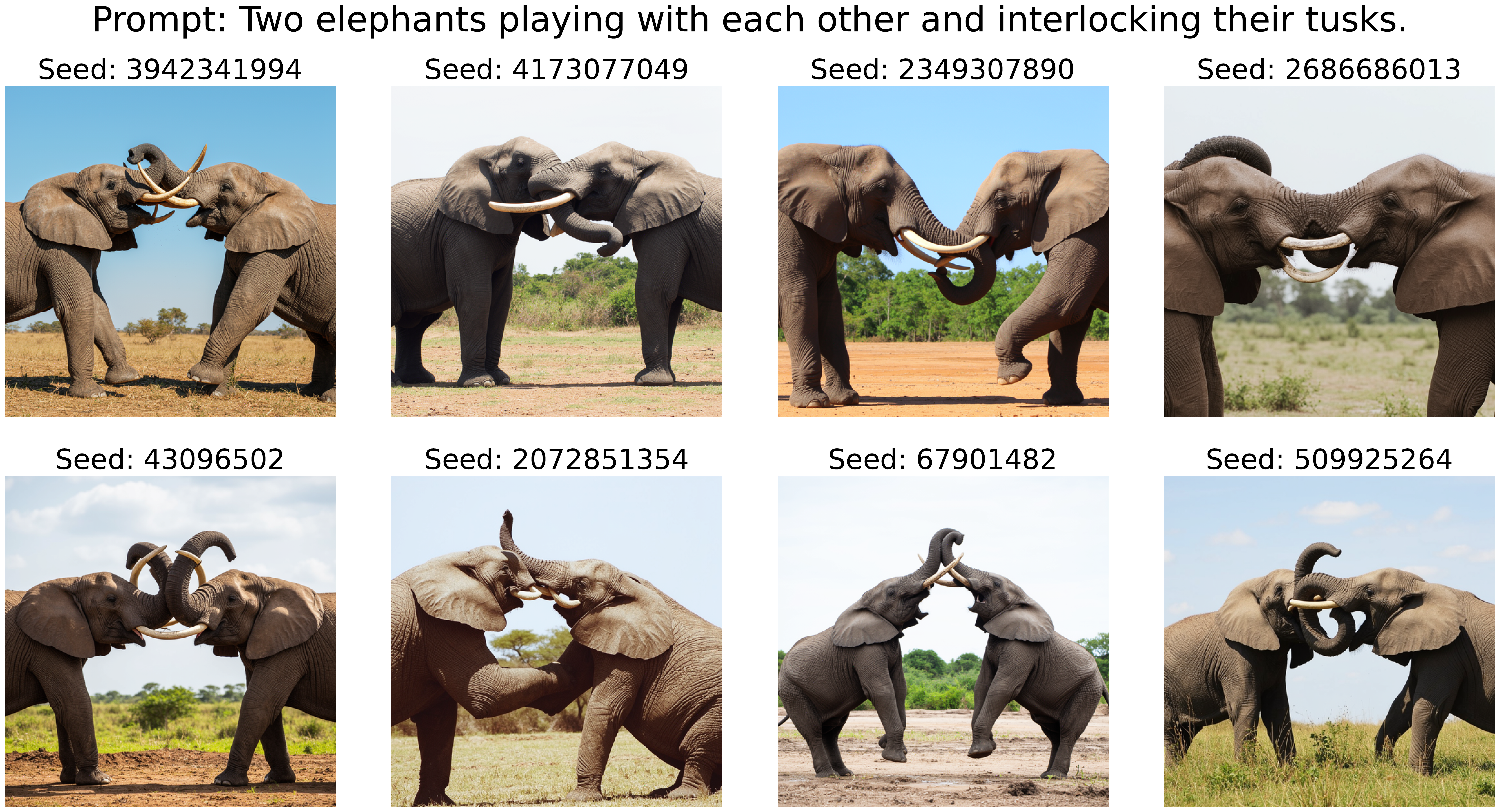}}
    \caption{\textbf{Compositional failure example (1/2).} The two elephants' heads occasionally fuse into a single shared head (e.g., top-right image), a compositional artifact present in the pre-trained FLUX.2-9B-base model prior to any fine-tuning.}
    \label{fig:compositional_failure_1}
\end{figure}

\begin{figure}[t]
    \centerline{\includegraphics[width=\linewidth]{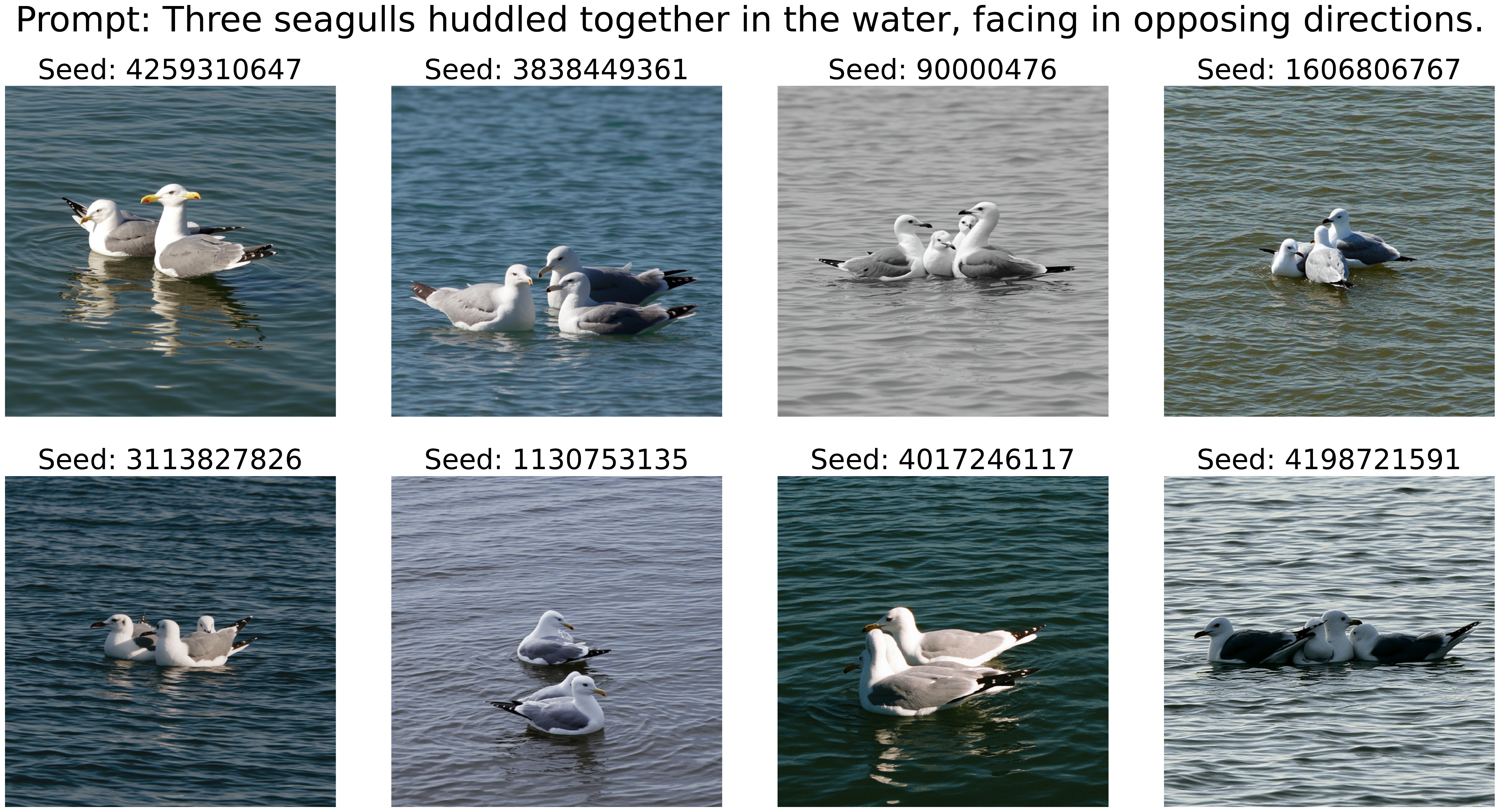}}
    \caption{\textbf{Compositional failure example (2/2).} Across samples, the seagull is occasionally rendered with two beaks pointing in different directions (e.g., top-left image), or with multiple heads (e.g., the two bottom-right images), a compositional artifact already present in the pre-trained FLUX.2-9B-base model prior to any fine-tuning.}
    \label{fig:compositional_failure_2}
\end{figure}

As noted in Sec.~\ref{sec:real_world_results}, compositional artifacts (e.g., merged or duplicated anatomical features) occasionally occur in generated images of our fine-tuned FLUX.2-9B-base model (Fig.~\ref{fig:qualitative_real_world_results}). However, as can be seen in Fig.~\ref{fig:compositional_failure_1} and Fig.~\ref{fig:compositional_failure_2}, such failure modes also occur in images generated using the \textit{pre-trained} FLUX.2-9B-base model prior to any fine-tuning. The images were generated using the standard FLUX.2 sampling configuration (same as in Sec.~\ref{sec:real_world_results}), demonstrating that compositional failures such as merged elephant heads (Fig.~\ref{fig:qualitative_real_world_results}, row 4) are already present in the base model rather than being introduced by our fine-tuning procedure. Each figure shows 8 samples (random integer seeds) for the fixed prompt given in the figure title, generated at $512\times512$ resolution with a guidance scale of 4.0 and 50 inference steps.

